\pdfoutput=1
\documentclass[11pt]{article}

\usepackage[preprint]{acl}

\usepackage{times}
\usepackage{latexsym}

\usepackage[T1]{fontenc}
\usepackage[utf8]{inputenc}

\usepackage{microtype}

\usepackage{inconsolata}

\usepackage{booktabs}
\usepackage{graphicx}
\usepackage{float}
\usepackage{caption}

\title{TartuNLP @ SIGTYP 2024 Shared Task: Adapting XLM-RoBERTa for Ancient and Historical Languages}

\author{Aleksei Dorkin \\
  Institute of Computer Science \\ University of Tartu \\
  {\tt aleksei.dorkin@ut.ee} \\\And
  Kairit Sirts \\
  Institute of Computer Science \\ University of Tartu \\
  {\tt kairit.sirts@ut.ee} \\}

\begin{document}
\maketitle
\begin{abstract}
We present our submission to the unconstrained subtask of the SIGTYP 2024 Shared Task on Word Embedding Evaluation for Ancient and Historical Languages for morphological annotation, POS-tagging, lemmatization, character- and word-level gap-filling.
We developed a simple, uniform, and computationally lightweight approach  based on the adapters framework using parameter-efficient fine-tuning. 
We applied the same adapter-based approach uniformly to all tasks and 16 languages by fine-tuning stacked language- and task-specific adapters. 
Our submission obtained an overall second place out of three submissions, with the first place in word-level gap-filling. 
Our results show the feasibility of adapting language models pre-trained on modern languages to historical and ancient languages via adapter training.

\end{abstract}

\section{Introduction}

The application of natural language processing techniques and pre-trained language models to analysis of ancient and historical languages is a compelling subject of research that has been so far overlooked.
While there exist a number of benchmarks, such as GLUE \cite{wang2018glue}, SuperGLUE \cite{wang2019superglue}, or XGLUE \cite{liang2020xglue}, for evaluating the quality of embeddings and language models for modern languages, such benchmarks are lacking for ancient and historical languages. Thus, the SIGTYPE 2024 Shared Task on Word Embedding Evaluation for Ancient and Historical Languages contributes to filling this gap.

In the current transformer-based language models paradigm, one of the common approaches to solving the tasks present in such benchmarks is to use the task data to fine-tune an encoder transformer model. The approach was first in introduced in \citet{devlin-etal-2019-bert} and was shown to yield superior results compared to the alternatives on the GLUE benchmark. Additionally, the proposed pre-training method is very similar to the word-level gap-filling problem in the shared task. Consequently, the model can be applied to solving the problem directly. This motivates our choice to use a transformer model in the shared task.

Large pre-trained language models, however, are predominantly trained on corpora of modern languages, with few exceptions such as LatinBERT~\cite{bamman2020latin}. Ancient and historical languages generally lack sufficient data to perform full pre-training of large language models or to continue training from a checkpoint trained on some different language. Full fine-tuning  of modern language models on a relatively small amount data in an ancient/historical language might lead to overfitting and catastrophic forgetting. These considerations are also common in context of other low resource tasks or domains. 

Several approaches have been proposed to alleviate these issues. For instance, the supplementary training approach proposed by \citet{phang2018sentence} involves first fine-tuning a pre-trained model on an intermediary labeled task with abundant data, and then on the target task which may have limited data. This approach showed gains over simply fine-tuning on the target task. 
\citet{pfeiffer2020mad} developed a cross-lingual transfer-learning approach based on the adapters framework for parameter efficient fine-tuning of language models~\cite{houlsby2019parameter, bapna2019simple}. Their approach involves fine-tuning a language adapter and a task adapter stacked on top of each other. This adapter-based method is expanded by \citet{pfeiffer2020unks} by adopting a custom tokenizer and an embedding layer. Several ways of initializing the new embedding layers are compared on underrepresented modern languages. Since ancient and historical languages are underrepresented, the same techniques should be applicable in this context as well.

\begin{figure*}[h]
  \centering
  \includegraphics[width=0.8\textwidth]{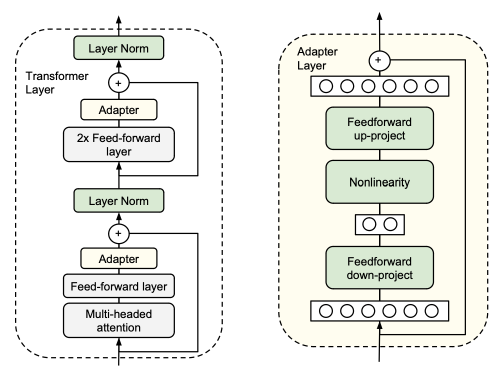}
  \caption{An illustration of the Bottleneck Adapter from \cite{houlsby2019parameter}. The left side demonstrates how a bottleneck adapter is added to a single transformer layer, while the structure of an individual adapter layer is on the right. Only elements in green are trained, while the rest remains frozen.}
  \label{fig:bn_adapter}
\end{figure*}

A useful feature of both the supplementary training approach of \citet{phang2018sentence} and the adapter-based training of \citet{pfeiffer2020mad} is that they provide a uniform framework that can be applied to different languages and tasks in a similar manner. While previous related works mainly focused on modern languages, we aim to assess the feasibility of this unified approach for ancient and historical languages. 
 
Our submission to the SIGTYP 2024 Shared Task on Ancient on Word Embedding Evaluation for Ancient and Historical Languages adopts the methods described by \citet{pfeiffer2020mad, pfeiffer2020unks} by stacking fine-tuned language and task adapters, and customizing the tokenizer and the embedding layers.
Our system is implemented as a unified framework, where various pre-trained models, languages, and tasks can be plugged in.
The system was evaluated on POS tagging, morphological annotation, lemmatization, and filling-in both word-level and character-level gaps for 16 ancient and historical languages. 
We participated in the unconstrained subtask, which allowed using any additional resources, such as pre-trained language models. Out of 3 participants on the leaderboard we took a close second place, with the first place on the word gap-filling task.  
Our main contribution is showing that by adopting the parameter-efficient adapter training methodology, the large language models pre-trained on modern languages are applicable also to low-resource ancient and historical languages.

\section{Adapters}

There exist a number of approaches to parameter efficient fine-tuning. In our submission to the shared task the focus is on adapters exclusively.

\citet{houlsby2019parameter} propose a parameter efficient fine-tuning strategy that involves injection of a number of additional trainable layers into the original architecture. The architecture of the proposed strategy is illustrated on Figure \ref{fig:bn_adapter}. In each transformer layer an adapter layer is added twice: after the attention sub-layer and after the feedforward sub-layer. To limit the number of trainable parameters, the authors propose a bottleneck architecture of adapter layers: the adapter first projects the input into a smaller dimension, applies non-linearity, then projects back to the original dimension. This is known as the bottleneck adapter.

\begin{figure}
  \centering
  \includegraphics[width=0.3\textwidth]{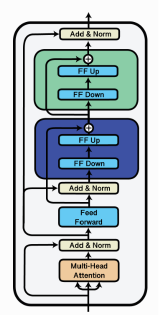}
  \caption{An illustration of an adapter stack as presented on the AdapterHub documentation page\protect\footnotemark. Blue and green blocks represent different adapter layers stacked on top of each other.}
  \label{fig:adapter_stack}
\end{figure}

\citet{pfeiffer2020mad} extend on the strategy in context of cross-lingual transfer learning in two ways. First, the adapter stack technique is introduced (illustrated on Figure \ref{fig:adapter_stack}), which is primarily used to separate language adaptation training and task-specific training. In the technique a language adapter is first trained, then a task-specific adapter is stacked on top of it and trained (while the language adapter is frozen). Secondly, the bottleneck adapter is expanded upon to include the embedding layer as well---the invertible adapter is introduced that transforms both input and output token representations to further improve language adaptation.

\begin{table*}[h]
\centering
\begin{tabular}{l|cccc}
\toprule
\textbf{Language}          & \textbf{Code} & \textbf{Train Sentences} & \textbf{Valid Sentences} & \textbf{Test Sentences} \\ 
\midrule
Ancient Greek              & grc           & 24,800           & 3,100            & 3,101           \\ 
Ancient Hebrew             & hbo           & 1,263            & 158              & 158             \\ 
Classical Chinese          & lzh           & 68,991           & 8,624            & 8,624           \\ 
Coptic                     & cop           & 1,730            & 216              & 217             \\ 
Gothic                     & got           & 4,320            & 540              & 541             \\ 
Medieval Icelandic         & isl           & 21,820           & 2,728            & 2,728           \\ 
Classical and Late Latin   & lat           & 16,769           & 2,096            & 2,097           \\ 
Medieval Latin             & latm          & 30,176           & 3,772            & 3,773           \\ 
Old Church Slavonic        & chu           & 18,102           & 2,263            & 2,263           \\ 
Old East Slavic            & orv           & 24,788           & 3,098            & 3,099           \\ 
Old French                 & fro           & 3,113            & 389              & 390             \\ 
Vedic Sanskrit             & san           & 3,197            & 400              & 400             \\ 
Old Hungarian              & ohu           & 21,346           & 2,668            & 2,669           \\ 
Old Irish                  & sga           & 8,748            & 1,093            & 1,094           \\ 
Middle Irish               & mga           & 14,308           & 1,789            & 1,789           \\ 
Early Modern Irish         & ghc           & 24,440           & 3,055            & 3,056           \\
\bottomrule
\end{tabular}
\caption{Language statistics in the Shared Task.}
\label{tab:lang_stats}
\end{table*}

\section{Methodology}

Our approach in this work is based on training language adapters and task specific adapters for XLM-RoBERTa~\cite{conneau2020unsupervised}---the multilingual variant of RoBERTa~\cite{liu2019roberta} trained on 100 different languages. Since many of the languages in the shared task have related modern languages in the training data of XLM-RoBERTa, or are even included themselves (such as Latin), we expect to benefit from knowledge transfer.

\subsection{Data}
The dataset provided by the organizers is a compilation of various resources
~\cite{
stgall2017,
wurzburg2018,
celt1997,
mgtsz2018,
11234/1-5150,
csnag,
simon2014}
and comprises 16 languages in total spanning several historical epochs and upper bounded by 1700 CE. The information on the languages in the dataset is presented in Table \ref{tab:lang_stats}.

\footnotetext{\url{https://docs.adapterhub.ml/adapter_composition.html}}

\subsection{Adapter Training}

Full fine-tuning of language models has several difficulties such as the possibility of catastrophic forgetting and the necessity to train and maintain a full copy of the model weights for each individual task. When applying the technique first proposed in \cite{phang2018sentence} in a multi-lingual and multi-task setting, the scale of the mentioned disadvantages is magnified. Training and maintaining a separate copy of the model for each combination of language, target task, and intermediate task can become computationally expensive. Meanwhile, each copy of the model is narrowly specialized in a single task, with generalization capabilities being potentially limited. This highlights the practicality of parameter efficient fine-tuning in comparison.

The overall approach is, in general, the same for every language. First, we train a language adapter for every language individually. A language adapter is comprised of a bottleneck adapter and masked language modeling prediction head. Accordingly, the training objective is masked language modeling. This is based on the assumption that a significant portion of the model's parameters is not actually language-specific, and thus does not need changing. We simply need to adapt the model to a new language. The language adapter is trained for 10 epochs regardless of the size of the data. The unmasked part of the training data for the word-filling task is used, while the masked part is not utilized for training. Secondly, we train task-specific adapters for each language. In this setup we use the adapter stack: the language adapter is loaded, but frozen, and we add a task-specific adapter on top of it, and train said adapter. We have two tasks per language: morphological tagging, which combines both POS-tagging and morphological annotation, and lemmatization, which is also implemented as sequence tagging problem. Similarly to language modeling, in both tasks the adapters are trained for 10 epochs.

\subsection{Custom tokenizers and embeddings}

Some of the languages in the shared task are not covered by the model's tokenizer. In other words, a significant portion of the tokenizer's output includes <unk> tokens. This is possible because XLM-RoBERTa's employs subword tokenization~\cite{sennrich-etal-2016-neural}. A subword tokenizer is a trained model that learns a vocabulary of a certain size. The trained model transforms text input into individual tokens by looking for the longest character sequences in the text that are present in its vocabulary. When attempting to tokenize a text that includes symbols that did not appear in the training data (and consequently, are not in the tokenizer's vocabulary), each unrecognized symbol is replaced by the <unk> token.

For each language not covered by the model's tokenizer a new tokenizer is initialized. In addition to that, a new embedding layer is also initialized. The initial weights are copied from the original embeddings for tokens that overlap with the multilingual tokenizer.

To decide whether the model requires a custom tokenizer, two checks are made:
\begin{enumerate}
    \item The percentage of unknown tokens > 5\%
    \item <unk> token is in top 10 by frequency
\end{enumerate}
If either of these conditions is true, a custom tokenizer is created. Refer to Table \ref{tab:token_stats} for detailed statistics. Based on these conditions, 5 languages need a separate tokenizer: Old Church Slavonic, Coptic, Classical Chinese, Old Hungarian, and Old East Slavic.
When selecting the vocabulary size, the aim was to have as many full words as possible in the vocabulary, but at the same time allow for morphological variation. Thus, the number of items in the vocabulary should be neither too high nor too low. For that reason, we selected the size to be 3000 for all languages, except Classical Chinese. For Classical Chinese the size was insufficient, so it was increased to 10000.

\citet{pfeiffer2020unks} suggest several possible options of initializing the weights of the custom embedding layer corresponding to the new tokenizer. These include random initialization, copying the weights of tokens that overlap between multilingual and language-specific tokenizers, as well as a matrix factorization-based approach that aims to identify latent semantic concepts in the original embedding matrix that are useful for transfer. It is noted, however, that the latter approach shows less gains when used with languages with underrepresented scripts. For this reason and for the sake of simplicity, with settled on using the lexical overlap approach for all affected languages.

\begin{table}[h!]
\centering
\begin{tabular}{lllll}
\toprule
\textbf{lang} & \textbf{\# tokens} & \textbf{\% unknown} & \textbf{<unk> rank} \\
\midrule
chu & 495612  & 15.31\% & 1 \\
cop & 39771 & 45.61\% & 2 \\ 
fro & 55448  & 0.00\% & - \\ 
ghc & 1282852 & 0.00\% & - \\ 
got & 98874 & 1.05\% & - \\ 
grc & 1079457 & 0.47\% & - \\ 
hbo & 124063  & 0.36\% & - \\ 
isl & 688580  & 0.00\% & - \\ 
lat & 296770 & 0.00\% & - \\
latm & 897209  & 0.00\% & - \\ 
lzh & 408259  & 1.40\% & 4 \\ 
mga & 492894  & 0.00\% & - \\
ohu & 352811  & 5.05\% & 1 \\
orv & 592890  & 5.00\% & 2 \\ 
san & 59349  & 0.01\% & - \\ 
sga & 190711  & 0.00\% & - \\
\bottomrule
\end{tabular}
\caption{Tokenizer coverage statistics. <unk> ranks not present in top 10 are not reported.}
\label{tab:token_stats}
\end{table}

\begin{figure*}
  \centering
  \includegraphics[width=0.8\textwidth]{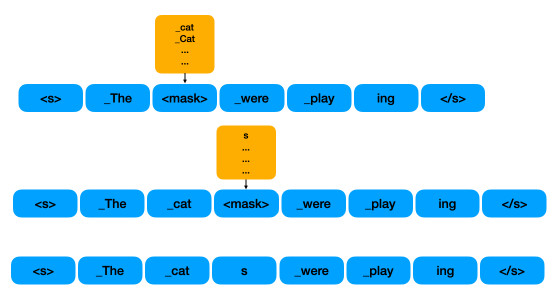}
  \caption{A schematic illustration of the decoding process for word-level mask filling. Blue boxes represent current tokens in the sentence, while orange boxes represent the probability distribution of tokens at the position of the mask token. The upper half represents the first step of decoding. We start with predicting the most likely replacement for the leftmost masked token that starts with \_ symbol representing the beginning of a new word. Then, we replace the mask with that token and append a new mask token to the right of it, as represented in the middle part. We predict the most likely replacement for the new mask token. If it starts with \_ symbol, we discard the mask token, consider the word predicted, and move to the next masked word if it's present. Conversely, if there's no \_ in the predicted token, we append it to the previously predicted token. We repeat this process $k$ times, or until we encounter a token starting with \_. $k$ is a hyperparameter that may be tuned, however increasing $k$ increases the decoding time significantly. For this reason we set $k$ to 1 for all languages. At the bottom of the figure the final result with no mask tokens is demonstrated. Note that this description is specific to XLM-RoBERTa tokenizer, other model's tokenizers may have different behaviour.}
  \label{fig:cats}
\end{figure*}

\subsection{Tasks}\label{subsection:tasks}

\paragraph{Filling Word-level Gaps} Having trained a masked language modeling adapter for each language, we already have a suitable model to perform word-level gap filling. However, the fact XLM-RoBERTa, as well as most modern language models, is based on subword tokens rather than full words becomes a significant hurdle. The reason is the we can not know in advance whether the target word we want to predict is comprised of one or more subword units, and that makes the decoding more complicated. To address this issue, for each sentence with masked words in it, we follow the steps (see Figure~\ref{fig:cats}):
\begin{enumerate}
    \item Locate the first <mask> token
    \item Use the model's prediction to determine the most likely token that starts with the whitespace prefix
    \item Replace the <mask> token with the prediction, and look ahead up to $k$ steps: does the next token start with the whitespace token? If it does, we break the cycle, and continue processing the remaining <mask> tokens in the sentence, otherwise we append the next predicted token to our previous prediction, and repeat the look ahead.
\end{enumerate}

For Classical Chinese, the process is simplified, and we simply predicted the most likely token.

\paragraph{Filling Character-level Gaps} The approach to character-level mask filling is purely algorithmic. We start with building a vocabulary of masked word candidates. We consider a sequence of characters with no whitespaces to be a suitable candidate. The vocabulary is populated with such candidates. For each masked word a look up among all the words of the same length as the masked word is performed: a regular expression built from the masked word by replacing "[\_]" with "." is matched against these words. If there are matches, up to the first 3 matches are returned. For each match character replacements are extracted from each candidate. If there are no matches, and the masked word contains only one "[\_]", the masked word is split in two parts, and each part is matched against the dictionary. If there is at least one match, the whitespace is returned as the replacement.

The described algorithmic method could be extended in two ways:
\begin{enumerate}
    \item Multiple candidates could be reranked using the trained language model
    \item If there are no candidates, a suitable word could be generated using the trained model, similarly to the word-level gap-filling task
\end{enumerate}

However, the extensions were not developed for the shared task for the following reasons. Each sentence may contain multiple masked words, and in addition the whitespace symbol may be masked as well. This significantly increases the solution's complexity, yet the magnitude of the benefit is unclear.

Finally, for Classical Chinese the following approach was applied. First, the symbols in the original dataset were decomposed into the lowest possible forms which are mostly strokes and small indivisible units using the Hanzipy library\footnote{\url{https://github.com/Synkied/hanzipy}}, then a masked language model adapter was trained on that decomposed data. The model achieved relatively high masked language modeling evaluation accuracy---about 40\%---on the validation set, as measured the by the Trainer class in the transformers library. The trained model was then used to predict the most likely replacement for each masked symbol. The ground truth data provided by the organizers, however, remained in the original composed form. This proved to be a significant problem. An attempt was made to compose the symbols back into their original form algorithmically, since the library used for decomposition doesn't have the option to reconstruct the original symbol from its constituents, to our best knowledge. However, this attempt was unsuccessful given that our submission attained score of 0 on the test set.

\paragraph{POS-tagging \& Full Morphological Annotation} We frame both POS-tagging and full morphological annotation as a single token classification task. For each token we collect all available morphological features and the POS tag from the annotation. Then we concatenate them as a single string, which is then used as the class to predict. Then we employ the adapter stack technique: we load the corresponding language adapter, and create a task specific adapter with token classification prediction head. We only train the task specific adapter. For inference we decompose the predicted class strings back into individual tags. The approach is the same for every language.

\paragraph{Lemmatization} Similarly to the previous task, we frame lemmatization as token classification task. To achieve that we generate transformation rules for each lemma/form pair based on the technique first introduced in \cite{straka-2018-udpipe} and expanded upon in \cite{dorkin2023comparison}. The rules are comprised of individual edits that need to be made to transform the given form into its lemma. The edits are represented as a single string. We use that string as the label to predict. Consequently, inference is done in two steps: first the rule for each token is predicted, then that rule is applied to the form and the resulting lemma is returned. The approach is the same for all language except for Classical Chinese. For Classical Chinese we do not train a model, but rather use a simple dictionary look up which results in nearly 100\% accuracy.

\begin{table*}[ht]
\centering
\begin{tabular}{p{0.18\linewidth}|p{0.12\linewidth}p{0.10\linewidth}p{0.12\linewidth}p{0.12\linewidth}p{0.11\linewidth}p{0.1\linewidth}}
\toprule
{} & Overall results & POS-tagging & Lemmatization & Morphological analysis & Character gap-filling & Word gap-filling \\
\midrule
Ancient Greek            &             0.70  &         0.96 &           0.94 &                             0.97 &                                0.61 &                           0.03 \\
Ancient Hebrew           &             0.61 &         0.94 &           0.97 &                             0.95 &                                0.19 &                           0.00 \\
Classical Chinese        &             0.57 &         0.83 &           1.00 &                             0.89 &                                0.00 &                           0.10 \\
Coptic                   &             0.52 &         0.61 &           0.75 &                             0.75 &                                0.45 &                           0.02 \\
Gothic                   &             0.70 &         0.93 &           0.93 &                             0.92 &                                0.67 &                           0.03 \\
Medieval Icelandic       &             0.73 &         0.97 &           0.98 &                             0.96 &                                0.57 &                           0.17 \\
Classical and Late Latin &             0.73 &         0.96 &           0.97 &                             0.96 &                                0.66 &                           0.11 \\
Medieval Latin           &             0.76 &         0.99 &           0.99 &                             0.99 &                                0.70 &                           0.14 \\
Old Church Slavonic      &             0.50 &         0.66 &           0.60 &                             0.67 &                                0.54 &                           0.02 \\
Old East Slavic          &             0.56 &         0.76 &           0.69 &                             0.80 &                                0.48 &                           0.06 \\
Old French               &             0.69 &         0.95 &           0.92 &                             0.98 &                                0.52 &                           0.07 \\

Vedic Sanskrit           &             0.66 &         0.84 &           0.88 &                             0.86 &                                0.65 &                           0.05 \\
Old Hungarian            &             0.52 &         0.75 &           0.63 &                             0.76 &                                0.46 &                           0.00 \\
Old Irish                &             0.19 &         - &           - &                             - &                                0.35 &                           0.03 \\
Middle Irish             &             0.22 &         - &           - &                             - &                                0.39 &                           0.04 \\
Early Modern Irish       &             0.28 &         - &           - &                             - &                                0.50 &                           0.06 \\
\bottomrule
\end{tabular}
\caption{Results of our submission on the leadearboard. For the variants of Irish, training and test data was only provided for gap-filling tasks. }
\label{tab:results}
\end{table*}

\subsection{Technical Implementation Details}

The implementation is primarily based on the Adapters\footnote{\url{https://github.com/adapter-hub/adapters}}~\cite{poth-etal-2023-adapters} library and the provided training script examples. The most significant change in the training scripts is the addition of custom embedding layer initialization. Some aspects of lemmatization as a token classification task were adopted from our previous work. For languages not utilizing custom tokenizers we employ the invertible bottleneck adapters~\cite{pfeiffer2020mad}, while for the languages that do require custom tokenizers we employ the regular bottleneck adapters~\cite{houlsby2019parameter}. The motivation is that when we have a custom embedding layer, there is no need to adapt it additionally. In all experiments we use the base version of XLM-RoBERTa due to time constraints, however we can well expect that the large version would show improved performance. The code and the instructions to reproduce are available on the project's GitHub repository\footnote{\url{https://github.com/slowwavesleep/ancient-lang-adapters/tree/sigtyp2024}}.

XLM-RoBERTa uses SentencePiece~\cite{kudo-richardson-2018-sentencepiece} for tokenization. SentencePiece is a trainable tokenization model limited by a token vocabulary of a specific size. To be able to process languages with scripts which are underrepresented by XLM-RoBERTa's tokenizer, we train a new tokenizer for each language in question. The new tokenizer has to remain compatible with original model---it has to retain all the special tokens and their indices. To achieve that we use the \texttt{train\_new\_from\_iterator} method of the Tokenizer class instance in HuggingFace transformers associated with model. We supply it with the same data we use for the masked language modeling training.

For training we used a single NVIDIA Tesla V100 GPU on the University's High Performance Cluster~\cite{https://doi.org/10.23673/ph6n-0144}. The training time for a single task (including masked language modeling) on average was about 10-20 minutes, ranging from 2 to 40 minutes depending on the amount of data, which is notably less than it would take for full fine-tuning.

Initially, an error was present in our code that resulted in our models not having trainable language modeling prediction heads. In other words, during the language adaptation stage the models retained the original masked language prediction head of XLM-RoBERTa, rather than training a new language specific prediction head. To our surprise, after fixing the error, the difference turned out to be quite negligible, except for languages that relied on custom embeddings. For these languages, the effect of the fixing the error was most noticeable on the word-level gap-filling task---before that the models were using the prediction head which was mismatched in size with the custom embeddings.

\section{Results}

According to the results on Table \ref{tab:results}, our approach seems to generally underperform on languages that require custom tokenizers and embeddings: Classical Chinese, Coptic, Old Church Slavonic, Old East Slavic, Old Hungarian. This can be explained by the amount of data being too limited to be able to train meaningful input representations, while the lexical overlap technique failed to provide benefit due to significant differences in their scripts. We hypothesize that a more sophisticated approach to embedding initialization and tokenizer training could result in considerable benefits.

We note a surprisingly strong performance of our algorithmic approach to character-level gap-filling with the exception of Classical Chinese where the approach is simply not applicable. We expect the extensions to the approach outlined in Section \ref{subsection:tasks} could improve the results further. However, in context of the shared task, we do not believe that the possible improvements would be worth the effort. Due to the whitespaces appearing as masked characters there is no way to know in advance how many individual words in the sentence in total have masked characters in them. In addition to that, a single word may contain several masked characters. The combination of these two factors makes improvements based on language model rescoring quite computationally expensive. For Classical Chinese, however, we hypothesise that the score would be considerably higher if the ground truth was also in the decomposed form, because to the best of our understanding, reconstructing the original characters from individual strokes is not a straightforward task.

Despite taking the first place in the word-level gap filling task, the absolute scores are still very low. This is to be expected, because the difficulty of the task is compounded by the presence of multiple masked words per sentence, as well the reliance on subword tokenization of our approach. Additionally, the implementation may not be suitable at all for certain languages due to differences in the writing systems. Finally, we surmise that due to the amount of training being quite small for the masked language modeling problem, we cannot expect to be able to correctly predict completely new words, however we can, in theory, predict previously unseen word forms due to subword tokenization. This means that experimenting with the $k$ parameter used to limit the number of times the look ahead for word continuation is performed in the word-level gap filling problem could improve the results somewhat.

The pattern-based approach to lemmatization also performs reasonably well in most languages. The approach is restricted by the set of patterns derived based on the training set. If the evaluation set contains many words for which a suitable pattern is missing, the system simply cannot make a correct prediction. However, we believe that this is not the problem here. In particular, we note that those languages which have lower scores on lemmatization, also perform consistently lower on other tasks. Thus, we suggest that the lower performance is rather related to insufficient data for training meaningful representations.
In addition to underpeforming on languages with underrepresented writing systems, we observe somewhat low performance on Vedic Sanskrit as well. One explanation could be that the usage of diacritics may negatively affect the character-based transformation rule generation; however this hypothesis requires further investigation.

Finally, our combined approach to POS-tagging and morphological analysis performs quite well on most of the languages. This is somewhat unexpected, because the model is limited in predicting only the combinations of part-of-speech and morphological features that are present in training data.

\section{Conclusion}

This paper described our solution to the SIGTYP 2024 Shared Task developed based on the adapters framework. 
The system is simple and uniform, and can be easily extended to other tasks and languages. 
For each language we trained a language adapter, and two task adapters, one for POS and morphological tagging and one for lemmatization. For languages with scripts underrepresented in the XLM-RoBERTa vocabulary, we additionally created custom tokenizers and embeddings.
One notable advantage of the adopted approach is its resource efficiency---adapting the model to a new language and task can be done in less than an hour.
As a future work, the system could be improved with more advanced adapter-based techniques such as adapter fusion~\cite{DBLP:journals/corr/abs-2005-00247} to leverage typological relatedness of languages.

\section*{Acknowledgments}

This research was supported by the Estonian Research Council Grant PSG721.

\bibliography{custom}

\end{document}